\documentclass[letterpaper, 10 pt, conference]{ieeeconf}

\IEEEoverridecommandlockouts                              

\usepackage{amsmath} 

\usepackage[ruled,vlined]{algorithm2e}
\usepackage{graphicx}
\usepackage{flushend}
\usepackage{censor}

\title{\LARGE \bf

VersualRL: Closed-Loop Verbal Reinforcement Learning with \\ Visual Execution Feedback for Task-Level Robot Planning

}

\author{%
    Dmitrii Plotnikov*,
    Iaroslav Kolomiets*,
    Dmitrii Maliukov*,
    Dmitrij Kosenkov*,\thanks{*These authors contributed equally to this work.} \\
    Daniia Zinniatullina,
    Artem Trandofilov,
    Georgii Gazaryan,
    Kirill Bogatikov,
    Timofei Kozlov, \\
    Mikhail Konenkov,
    Miguel Altamirano Cabrera,
    and Dzmitry Tsetserukou%
    \thanks{The authors are with the Intelligent Space Robotics Laboratory,
    Skolkovo Institute of Science and Technology. 
    \{dmitrii.plotnikov2, iaroslav.kolomiets, dmitrii.maliukov,
    dmitrij.kosenkov, daniia.zinniatulina, artem.trandofilov,
    georgii.gazaryan, kirill.bogatikov, timofei.kozlov, mikhail.konenkov, m.altamirano, d.tsetserukou\}@skoltech.ru}
}

\usepackage{comment}
\usepackage{listings}
\begin{document}

\bstctlcite{IEEEexample:BSTcontrol}

\maketitle
 \thispagestyle{empty}
\pagestyle{empty}

\begin{abstract}

We introduce VersualRL, a closed-loop framework for task-level robot planning that uses visual execution feedback to iteratively refine executable Behavior Trees through structured natural-language critique. VersualRL combines a vision-language model critic, which analyzes visual observations and Behavior Tree execution traces, with a large language model actor that performs discrete and interpretable policy updates. During physical deployment, both foundation models remain frozen, while the executed Behavior Tree is updated at the symbolic level without online gradient-based policy optimization. This enables transparent reasoning, explicit causal feedback, and human-interpretable policy evolution. We validate VersualRL on a real mobile robot performing a multi-stage navigation and manipulation task under execution uncertainty. Experimental results show that the framework supports explainable policy improvements, closed-loop adaptation to execution failures, and successful deployment on physical robotic systems.

\end{abstract}

\section{Introduction}

In real-world environments, mobile robots operate under significant execution uncertainty \cite{uncerta}. Variations in layouts, imperfect localization, and low-level control errors make the design of robust task-level policies challenging and time-consuming \cite{ProbRob}.
In practice, such policies are often implemented using Behavior Trees (BTs), which provide modularity and transparency but require significant manual tuning to handle unexpected failures \cite{bt1,bt2}.

Reinforcement Learning (RL) offers a principled framework for autonomous policy improvement through interaction and iteration. Recent Deep RL (DRL) and Multi-Agent Deep RL (MADRL) methods have demonstrated strong performance in robotic motion planning and task allocation \cite{AgilePilot, HIPPO-MAT}.  However, most RL approaches optimize policies in a sub-symbolic space, producing behaviors that are difficult to interpret or debug \cite{aisafety}. This lack of transparency limits their applicability for the real-world deployment of robotic systems.

Recent works \cite{do_as_i_say_llm, Code_as_policy_llm, prog_prompt_llm_task_plan, autogpt_llm, Shinn2023NeurIPS} have explored the use of Large Language Models (LLMs) for robotic task planning, leveraging their reasoning and generalization capabilities \cite{Yao2022ReAct, Madaan2023NeurIPS}. While promising, LLM-based planners typically operate in an open-loop manner and lack grounded feedback from the physical world. As a result, they struggle to reliably adapt to real-world execution failures.

\begin{figure}
    \centering
    \scalebox{1}[1.0]{%
        \includegraphics[width=\linewidth]{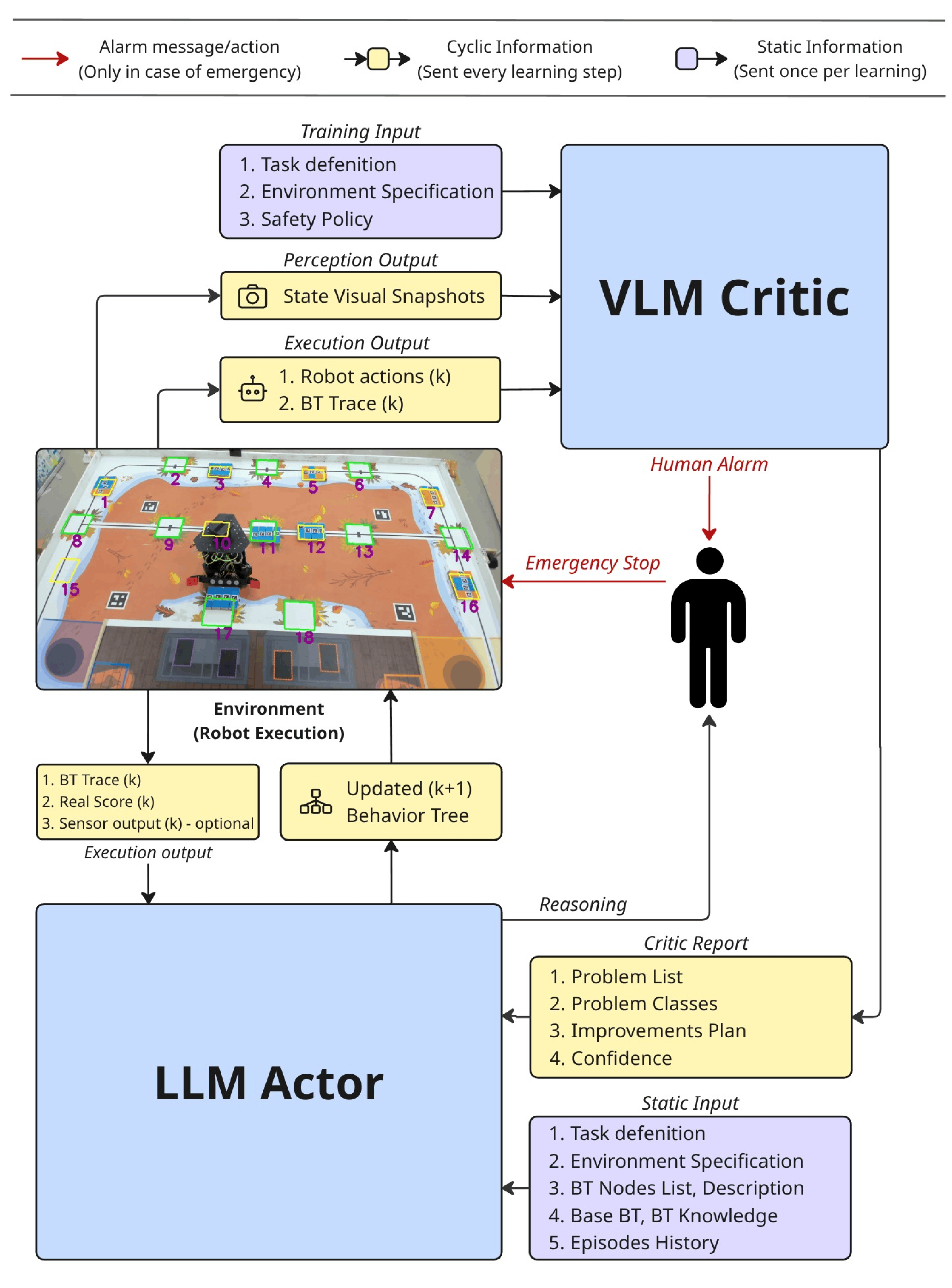}
    }
    \caption{Closed-loop VersualRL architecture. The VLM critic analyzes visual snapshots and BT execution traces to produce structured feedback, while the LLM actor updates the executable BT using critic feedback and the human-verified episode score $R$.}
    \vspace{-3mm}
    \label{fig:placeholder}
\end{figure}

To address these limitations, we propose VersualRL, a closed-loop framework for task-level robot planning that combines verbal policy refinement with visual execution feedback. In VersualRL, task policies are represented as executable Behavior Trees and iteratively refined through structured natural-language feedback. An LLM-based actor modifies the task plan, while a VLM critic analyzes robot operation using visual observations and BT execution traces \cite{vox_poser_llm}. Unlike Reflexion-style verbal self-refinement, \linebreak VersualRL grounds feedback in physical robot execution; unlike VLM-based reward learning such as RL-VLM-F, it directly edits an executable symbolic BT policy instead of learning a dense reward model or optimizing a neural policy.

\section{Related Work}

Recent advances in language and vision-language foundation models have enabled more flexible robotic perception, planning, and control.

\subsection{LLMs for Task-Level Planning and Code Generation}

Recent works employ large language models as high-level planners that translate natural-language instructions into executable code or symbolic task plans. Some systems \cite{GROP, Inner_monolog, Lykov2023LLMMARS,llm_again,overview_llm, Wang2024Voyager} combine LLM-based common-sense reasoning with task-and-motion planning, converting underspecified goals into commands that can be executed by classical planners. These approaches demonstrate strong semantic reasoning, but they typically operate in a one-shot or open-loop manner and lack systematic mechanisms for strategy refinement.

BTs provide an interpretable and modular representation for LLM-driven planning. Previous research showed that LLMs can synthesize structurally valid BTs from natural-language commands \cite{btgenbot,LLMBRAIn}. However, existing approaches focus primarily on initial plan generation and do not address post-deployment policy improvement.

\subsection{Vision–Language Models for Reward Learning and Evaluation}

Vision–language models have been widely explored for reward learning in robotics. Some methods \cite{RL-VLM,vlm_zeroshot_reward, Singh2025VARPRL, Dang2025CLIPMotion, Zeng2024ICML} derive preference-based rewards from visual comparisons, providing automated supervision for reinforcement learning. These approaches rely on gradient-based optimization in sub-symbolic policy spaces, limiting their interpretability. Other works employ VLMs for affordance extraction, scene understanding, and autonomous driving assistance \cite{VLMAuto}, primarily as perception modules rather than explicit execution critics.

\subsection{Critic-Based Policy Refinement}

Recent studies investigate VLMs \cite{affordance_rl, vlm_critic} and agentic RL loops \cite{AgenticRL} as behavioral critics or self-refining mechanisms that identify undesirable actions and support iterative refinement. However, prior methods typically evaluate or rank policies without directly modifying structured symbolic representations such as Behavior Trees.

In contrast, our work couples a VLM-based visual critic with an LLM actor that performs direct, interpretable updates of executable Behavior Trees. This enables closed-loop task-level policy refinement on a physical robot without gradient-based reward learning or simulation.

\section{VersualRL Framework for Task-Level Robotic Planning}


This section formalizes our framework and describes the actor–critic interaction, symbolic policy representation, and execution-driven refinement mechanism.

\subsection{Policy Representation}

Task policies are represented as executable BTs, which provide modularity and interpretability.

The actor modifies BT structure and parameters directly, operating on a library of predefined expert-designed nodes and subtrees.  
This allows policy updates to remain transparent, verifiable, and grounded in existing expert knowledge.
The initial BT is deliberately simple and does not encode the task strategy, serving only as a basis for iterative improvement.

\subsection{LLM Actor: Symbolic Policy Refinement}

The actor is an LLM that operates at the symbolic level, refining an existing BT based on structured feedback. To constrain the search space and ensure syntactic validity of generated policies, the actor is prompted with a structured context consisting of: (i) the task definition, (ii) an environment specification, (iii) a library of available BT nodes with their descriptions, and (iv) BT authoring knowledge (coding rules) together with the current (initial) BT. Importantly, the actor does not receive raw sensory streams and does not directly control the robot; instead, it proposes discrete, interpretable edits to the BT structure and parameters.

Optionally, the actor’s input interface can be augmented with additional structured, perception- or sensor-derived state estimates when available. We treat this as an ablation on actor observability and evaluate it in Section V by providing symbolic block color/orientation information to the actor in addition to critic feedback, demonstrating that such structured inputs can substantially improve learning stability and overall performance.

\subsection{VLM Critic: Visual Execution Feedback}

The critic is a VLM that observes task execution through visual snapshots and BT execution traces. Instead of estimating scalar rewards, it outputs structured natural-language feedback grounded in observable evidence. To support different stages of execution, the critic operates in three modes: \textit{Initial}, \textit{Intermediate}, and \textit{Final}.

In addition to the textual critique, the critic outputs two scalars: an alarm score $s \in [0,1]$ indicating issue severity, and a confidence $c \in [0,1]$ indicating certainty. The alarm score is supervised using an ordinal severity rubric. In \textit{Initial} mode we annotate $s{=}0.0$ for clean setup, $s \in [0.1,0.3]$ for minor imperfections, $s \in (0.3,0.5)$ for borderline cases, $s \in [0.5,0.7]$ for actionable problems, and $s>0.7$ for severe safety-critical issues. For \textit{Intermediate/Final} (Human Alarm) annotations we use the same severity semantics but a slightly more tolerant rubric to account for inevitable manipulation noise, while preserving $s \ge 0.5$ as actionable problems and $s>0.7$ as severe failures.

After each episode, a human evaluator verifies the physical task outcome and computes the human-verified episode score $R$ according to the fixed scoring rules in Table~\ref{tab:placeholder}. The score is based on predefined task conditions and is not predicted by the critic.

VersualRL operates in an episodic closed loop. At each episode, the current BT is executed on the robot and the critic analyzes observations and traces to produce feedback. If $s \ge 0.5$ or $c < 0.3$, a human operator is notified. The actor then updates the BT using the final critic feedback and the human-verified episode score $R$. The overall procedure is summarized in Algorithm~1.

\vspace*{-0mm}  

\begin{algorithm}[ht]
\caption{VersualRL Policy Refinement}
\label{alg:vrl}
\KwIn{Initial Behavior Tree $BT_0$, Actor $\mathcal{A}$, Critic $\mathcal{C}$}
\KwOut{Refined Behavior Tree $BT^*$}

$BT \leftarrow BT_0$\;
Initialize actor memory ${H} \leftarrow \emptyset$\;

\For{episode $= 1$ to $N$}{
    
    Reset critic episodic memory $M \leftarrow \emptyset$\;
    
    Capture initial image $I_0$\;
    $y_0 \leftarrow \mathcal{C}.\textsc{Initial}(I_0)$\;
    \If{$\textsc{Alarm}(y_0)$}{Notify human operator\;}
    
    \While{episode not terminated}{
        Execute next subtree of $BT$\;
        Capture image $I_t$ and BT execution trace $T_t$\;
        $y_t \leftarrow \mathcal{C}.\textsc{Intermediate}(I_t, T_t)$\;
        $M \leftarrow M \cup \{(I_t, T_t, y_t)\}$\;
        \If{$\textsc{Alarm}(y_t)$}{Pause execution and wait for operator intervention\;}
    }
    
    Capture final image $I_f$ and complete BT execution trace $T$\;
    ${F} \leftarrow \mathcal{C}.\textsc{Final}(I_f, T, M)$\;
    \If{$\textsc{Alarm}({F})$}{Notify human operator\;}
    
    Compute human-verified episode score $R$\;
    $BT \leftarrow \mathcal{A}(BT, {F}, R, {H})$\;
    
    ${H} \leftarrow {H} \cup \{{F}, R\}$\;
}

$BT^* \leftarrow BT$\;
\Return{$BT^*$}\;

\end{algorithm}
Here, $F$ denotes the final critic report, $R$ is the human-verified episode score computed according to Table~\ref{tab:placeholder}, and $H$ stores prior reports and scores.
\section{Experimental Setup}

\subsection{Robotic Platform}
\begin{figure}[!t]
    \vspace{2mm}
    \centering
    \includegraphics[width=0.8\linewidth]{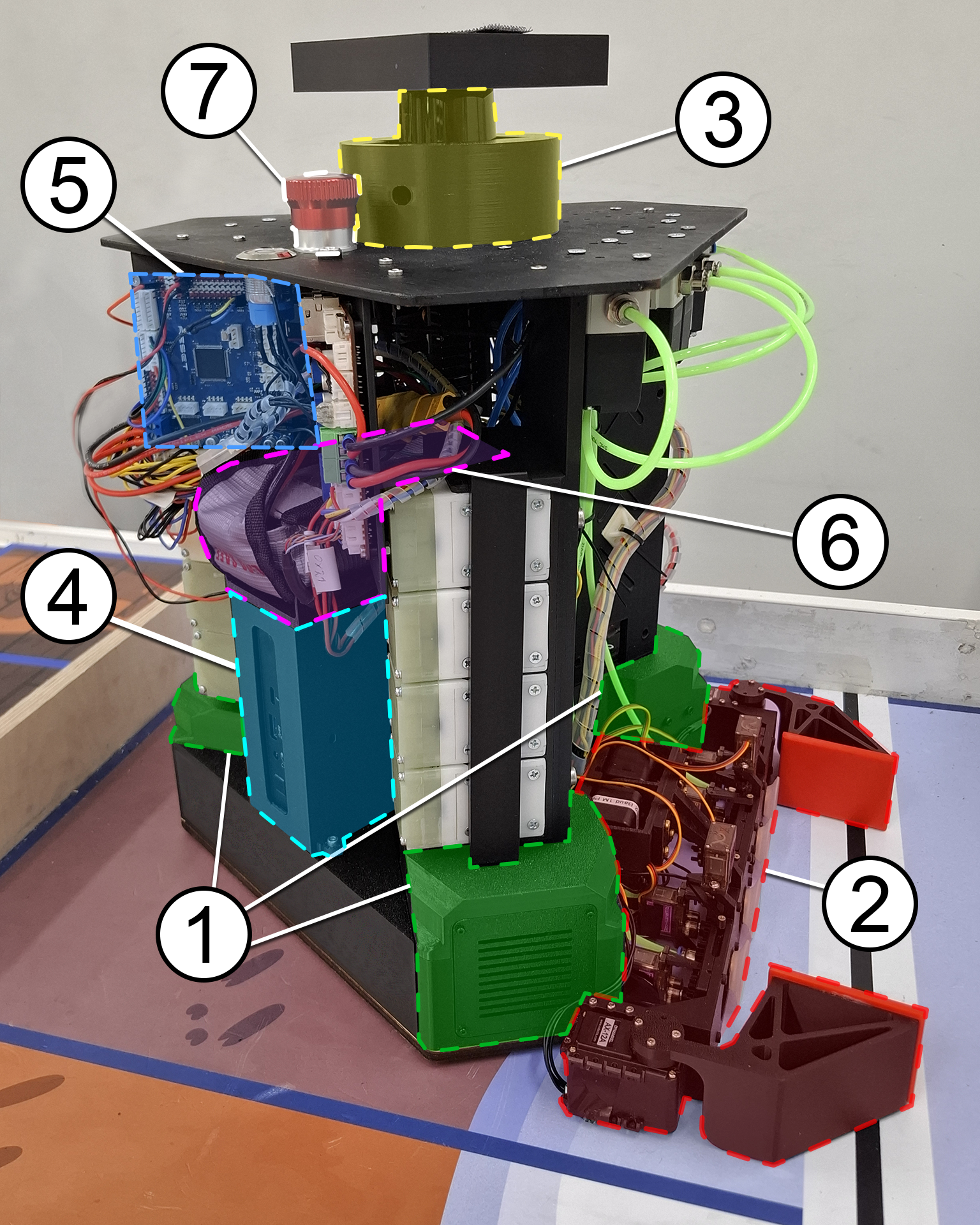}
    \caption{Hardware architecture of the mobile robot: (1) three-wheeled
    omnidirectional base, (2) custom servo-actuated gripper with an auxiliary
    vacuum channel, (3) LiDAR sensor, (4) onboard Intel NUC computer,
    (5) custom STM32-based low-level controller, (6) LiPo battery, and
    (7) emergency stop button.}
    \label{fig:robot}
    \vspace{-4mm}
\end{figure}

The experiments were conducted on a custom mobile manipulation robot designed to perform logistics tasks (Fig.~\ref{fig:robot}). 

The robot performs autonomous navigation between predefined zones and uses a servo-actuated gripper with an auxiliary vacuum channel to pick, rotate, and place blocks. The manipulation system can handle a fixed batch size of up to four blocks per transport cycle.

\subsection{Experimental Environment}

\begin{figure}[h]
    \centering
    \includegraphics[width=1\linewidth]{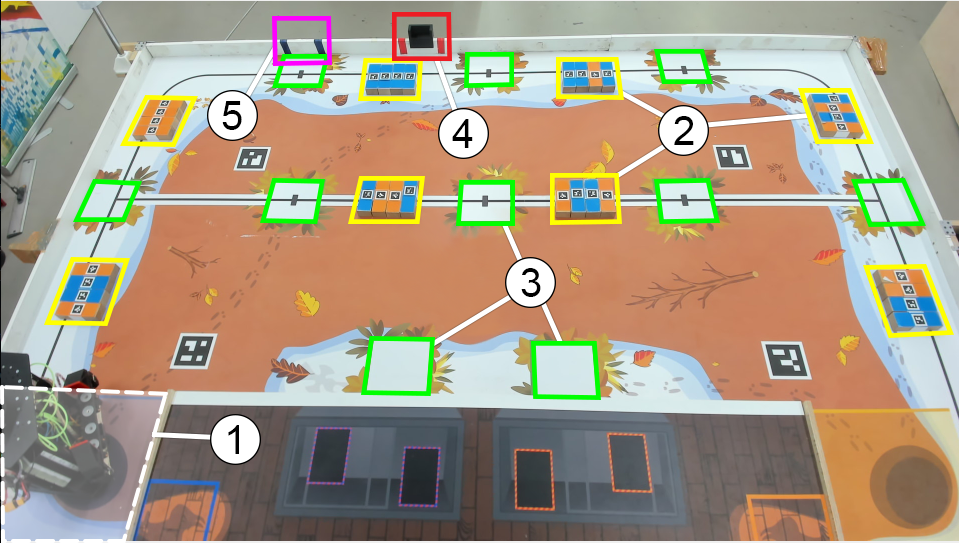}
    \caption{Layout of the experimental environment: (1) start/finish area, (2) load zones containing block sets, (3) designated unload zones, (4) initial position of the movable shelf, and (5) target position of the movable shelf.}
    \vspace{-3.0mm}
    \label{fig:field_start}
\end{figure}

We established certain specifications for the experimental field, including the existence of several load and unload zones, the designated starting area, and a movable object (shelf). For the purpose of experiments, we used the available robotics competition environment (Fig.~\ref{fig:field_start}). The geometry of the field provides spatially distributed pickup and delivery zones suitable for warehouse-style manipulation tasks.

We use wooden blocks of 150 mm $\times$ 50 mm $\times$ 30 mm. Two opposite large faces are painted orange and blue, while the remaining faces are unpainted. The color of the upward-facing side determines the orientation of the block. If the orange face is upward, the block is incorrectly oriented (upside down).

\subsection{Task Definition}
The experimental task represents a structured warehouse logistics scenario under a time constraint. The robot is required to transport blocks from load zones to unload zones. Blocks must be collected in batches of up to four units and placed such that each block is at least partially within an unload zone (i.e., any overlap/contact with the zone is sufficient). Each block must also be correctly oriented: the upward-facing colored side determines whether the placement is valid. In addition to block transportation, the robot must relocate a movable shelf along the boundary to the designated final side of the field. Successful relocation implicitly requires clearing adjacent zones, although this constraint is not explicitly encoded in the prompts. The episode is considered complete when the robot returns to the finish area.

\subsection{Critic Models and Fine-Tuning}

The Qwen2.5-VL-3B critic was adapted via parameter-efficient supervised
fine-tuning on 88 execution episodes (65 training and 23 validation),
split by recording session. Each sample pairs an annotated field image
and its BT execution trace with a target critique following the severity
rubric in Section~III-C. LoRA adapters were applied to the attention and
MLP projections of the 8-bit-quantized base model. The checkpoint with
the lowest validation loss, reached near epoch 16, was used in the
experiments. Full hyperparameters are reported in
Table~\ref{tab:finetune_hparams}.

Qwen2.5-VL-7B-Instruct and Gemini-3-Pro-Preview were evaluated without
task-specific fine-tuning. All critics remained frozen during episodic
policy refinement; only the Behavior Tree was updated.

\begin{table}[t]
    \vspace*{3mm}
    \centering
    \caption{Fine-tuning Hyperparameters for the Qwen2.5-VL-3B Critic}
    \label{tab:finetune_hparams}
    \scriptsize
    \begin{tabular}{|p{0.34\columnwidth}|p{0.56\columnwidth}|}
        \hline
        Parameter & Value\\
        \hline
        Base model & Qwen2.5-VL-3B-Instruct (8-bit, bitsandbytes)\\
        \hline
        LoRA rank $r$ / $\alpha$ & 16 / 16\\
        \hline
        LoRA dropout & 0.15\\
        \hline
        Target modules & Attention and MLP projections\\
        \hline
        Optimizer & AdamW (fused), $\beta=(0.9,0.999)$\\
        \hline
        Learning rate / schedule & $5\times10^{-5}$, linear, warmup ratio 0.15\\
        \hline
        Effective batch size & 8 (batch size 1, gradient accumulation 8)\\
        \hline
        Epochs & Up to 120 epochs\\
        \hline
        Precision & bf16, gradient checkpointing\\
        \hline
        Checkpoint selection & Lowest validation loss ($\approx$ epoch 16)\\
        \hline
        Hardware & 1$\times$ RTX 4090 24 GB, $\approx$7 h\\
        \hline
    \end{tabular}
    \vspace{-2mm}
\end{table}

We evaluate five distinct experimental configurations:

1. Without critic (BT execution trace + human-verified episode score $R$ only)

2. Qwen2.5-VL-7B Instruct

3. Qwen2.5-VL-3B Instruct (fine-tuned)

4. Gemini-3-Pro-Preview (MLLM as critic)

5. Gemini-3-Pro-Preview + block color information

Gemini-3-Pro-Preview was used as a closed-source VLM critic available at the time of data collection. The model was subsequently discontinued on March 9, 2026; therefore, the reported Gemini-based results correspond to the historical preview version used in our experiments and cannot be directly reproduced through the original endpoint.

Each configuration was evaluated on five field layouts with varying block
orientations using ten sequential episodes per layout. After each episode,
the actor updated the BT using critic feedback and the human-verified score
$R$.

\section{Experimental Results}

For each VLM, we report results averaged across all environment configurations. We analyze episodic learning behavior using four key metrics: human-verified episode score $R$, Issue Detection Recall, confidence stability, and human alarm rate. Together, these metrics characterize both the effectiveness of policy refinement and the reliability of visual feedback within VersualRL.

\subsection{Quantitative Learning Results}
\begin{figure}
    \vspace*{2mm}
    \centering
    \includegraphics[width=1\linewidth]{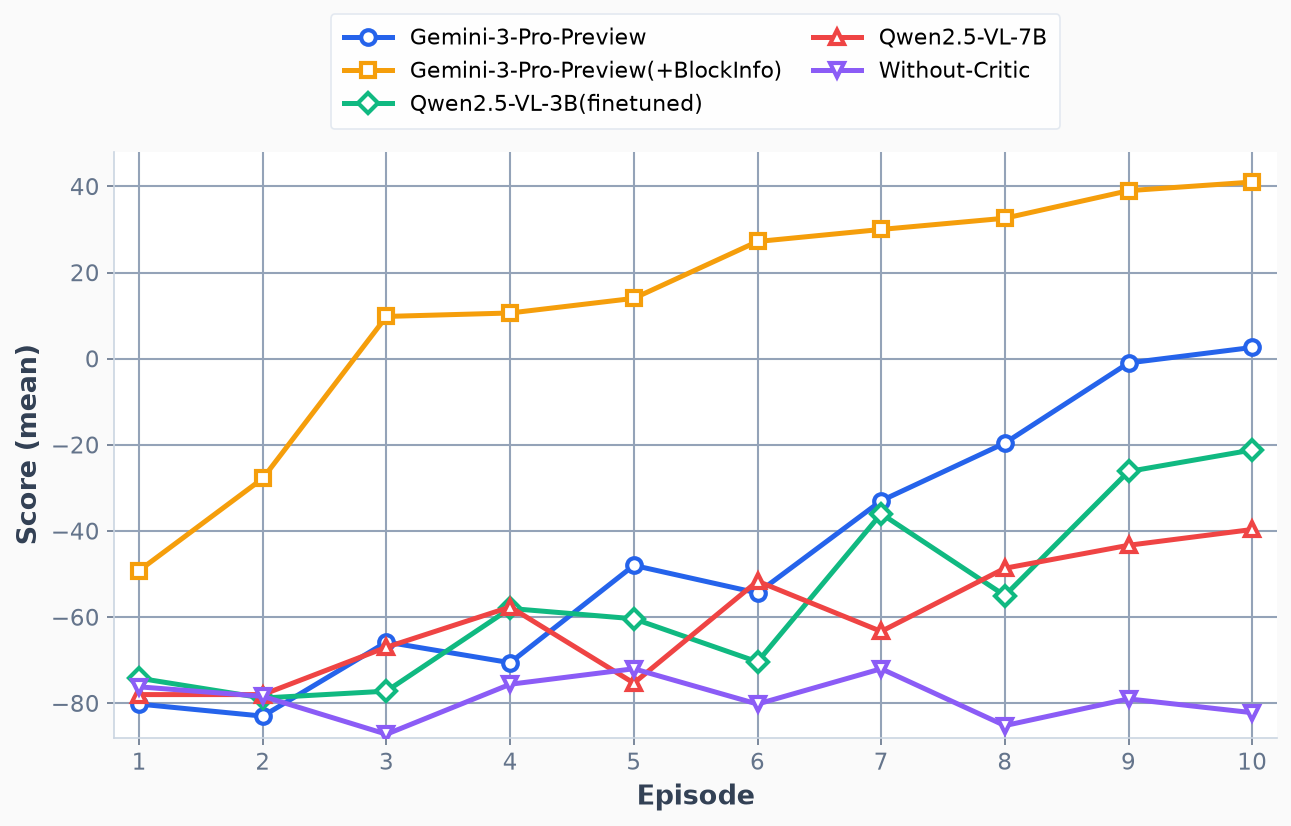}
    \caption{Mean human-verified episode score $R$ across environments.}
    \label{fig:score}
    \vspace{-3mm}
\end{figure}

\textit{Human-verified episode score.}
The human-verified episode score $R$ (Fig.~\ref{fig:score}) is computed according to a fixed rule-based evaluation function that reflects task completion quality. The scoring scheme, summarized in Table~\ref{tab:placeholder}, is designed to encourage structured logistics behavior and penalize operational errors.

\begin{table}[h]
    \centering
\caption{Rules for the Human-verified Episode Score $R$.}
\label{tab:placeholder}
    \begin{tabular}{|l|l|}\hline
         Condition& Score\\\hline
         Operating time penalty& $-$1 point per second\\\hline
 Additional timeout penalty ($t>150$ s)&$-$20 points\\\hline
         Block correctly placed& $+$10 points per block\\\hline
 Full batch of 4 correctly placed blocks&$+$10 bonus points\\\hline
 Block incorrectly placed&$-$5 points per block\\\hline
 Block completely outside any zone&$-$10 points per block\\\hline
 Correct final robot position&$+$20 points\\ \hline
    \end{tabular}

\end{table}

We first analyze the episodic dynamics of $R$, which directly reflect
task-level performance and policy refinement quality. The fine-tuned
Qwen2.5-VL-3B critic demonstrates more consistent improvement in $R$
than the larger Qwen2.5-VL-7B model. Despite its smaller size, the 3B
model benefits from task-specific fine-tuning on structured execution
feedback, producing more grounded and actionable critiques. In contrast,
the untuned 7B model exhibits slower improvement and higher variability,
indicating that domain adaptation is more important than model scale in
this setting.

Gemini-3-Pro-Preview converges faster and reaches higher final values of
$R$ than both Qwen variants, suggesting stronger visual reasoning and
feedback quality. Providing symbolic block color information further
improves learning stability: Gemini (+BlockInfo) achieves the highest
final performance and the most consistent convergence across episodes.
This result indicates that structured perceptual inputs reduce ambiguity
and support more precise strategy-level corrections.

The no-critic baseline shows substantially slower and less consistent
improvement. Without structured visual feedback, the actor relies only
on the scalar signal $R$, which indicates overall task performance but
does not reveal the causes of execution failures. Consequently, policy
refinement is less targeted and less stable than in the critic-guided
configurations.

\begin{figure}
    \vspace*{2mm}
    \centering
    \includegraphics[width=1\linewidth]{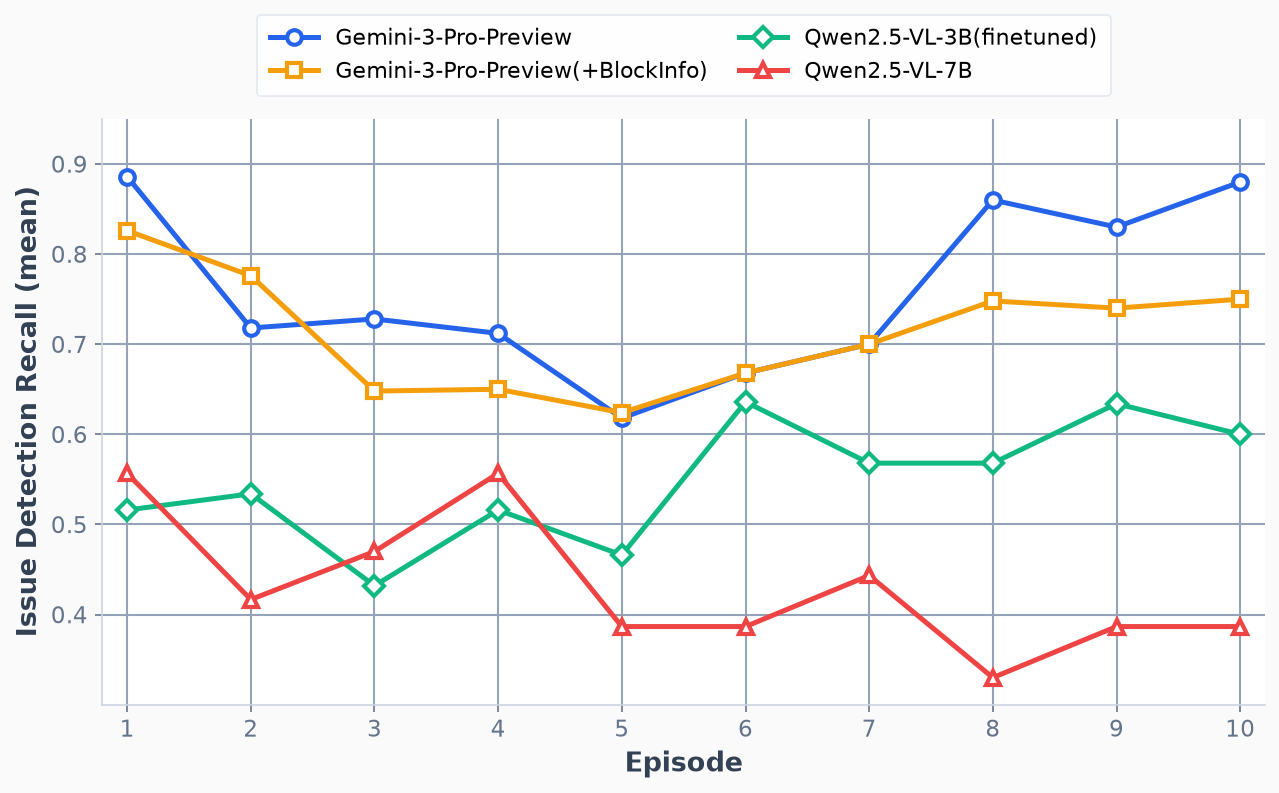}
    \caption{Mean Issue Detection Recall (IDR) across environments.}
    \label{fig:idr} 
\end{figure}

\textit{Issue Detection Recall.} 
Issue Detection Recall (IDR) reflects how reliably the critic detects ground-truth execution issues during task performance (Fig.~\ref{fig:idr}).

We quantify error detection performance using the Issue Detection Recall (IDR):

\begin{equation}
\mathrm{IDR}
=
\frac{N_{\mathrm{TP}}}
     {N_{\mathrm{TP}} + N_{\mathrm{FN}}},
\end{equation}
where $N_{\mathrm{TP}}$ is the number of ground-truth issues detected by the critic, and $N_{\mathrm{FN}}$ is the number of ground-truth issues missed by the critic.

Our fine-tuned Qwen2.5-VL-3B model demonstrates higher and more stable recall compared to the larger Qwen2.5-VL-7B model. Despite having fewer parameters, the domain-adapted 3B critic better captures task-specific failure modes, indicating that fine-tuning improves perceptual grounding and structured reasoning about execution errors.

The Qwen2.5-VL-7B model shows lower and more fluctuating recall across episodes, suggesting less consistent alignment between visual observations and task-relevant error categories.

Moving to Gemini-3-Pro-Preview, we observe substantially higher issue detection recall compared to both Qwen variants. The model demonstrates strong visual grounding and reliable identification of incorrectly placed or misoriented blocks across most episodes.

When additional symbolic block color information is provided, Gemini (+BlockInfo) does not always achieve higher IDR than the image-only Gemini critic. This is because IDR measures issue detection, not final task performance. BlockInfo mainly helps resolve color/orientation ambiguity and improves actor updates, but remaining errors may involve spatial placement, timing, or manipulation failures. Therefore, the BlockInfo variant can achieve better task scores despite slightly lower issue detection recall in some episodes.

\begin{figure}
    \vspace*{2mm}
    \centering
    \includegraphics[width=1\linewidth]{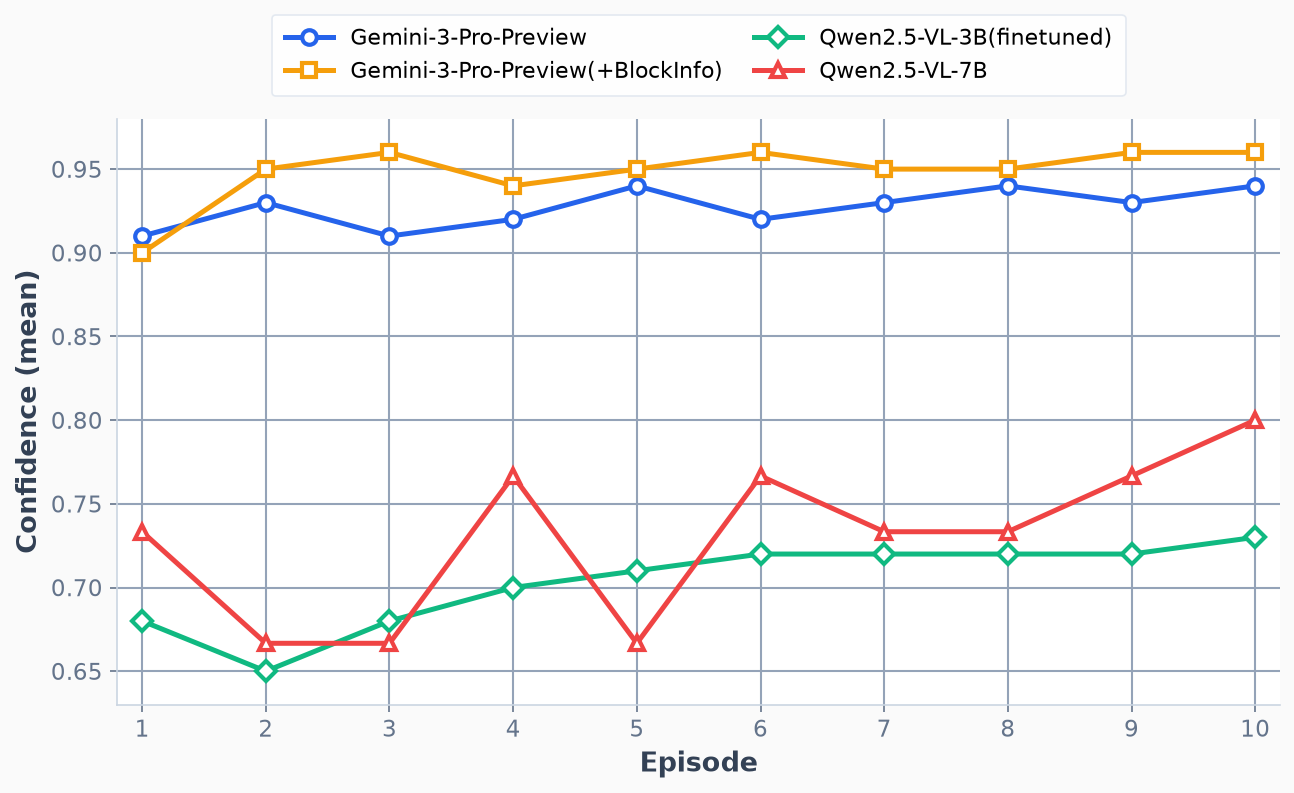}
    \caption{Mean Confidence across environments.}
    \label{fig:confidence}
\end{figure}

\textit{Confidence stability.} Confidence reflects how certain the critic is about its assessments during execution (Fig.~\ref{fig:confidence}). While recall measures detected issue coverage, confidence stability characterizes how consistently the critic reports its certainty across episodes.

The fine-tuned Qwen2.5-VL-3B model demonstrates more stable confidence levels across episodes compared to the larger Qwen2.5-VL-7B model. The 3B critic maintains moderate but consistent confidence, indicating more stable confidence after task-specific fine-tuning. In contrast, the 7B model exhibits higher variance in confidence, indicating less stable alignment between visual observations and internal reasoning.

The Gemini-3-Pro-Preview critic maintains generally high confidence throughout training, reflecting strong visual grounding capabilities. However, without structured block information, confidence occasionally remains high even in early episodes where planning errors are frequent.

When block color information is provided, Gemini (+BlockInfo) shows both high and more consistent confidence. The additional symbolic input appears to reduce perceptual ambiguity, leading to more stable confidence levels.  

\begin{figure}
    \vspace*{2mm}
    \centering
    \includegraphics[width=1\linewidth]{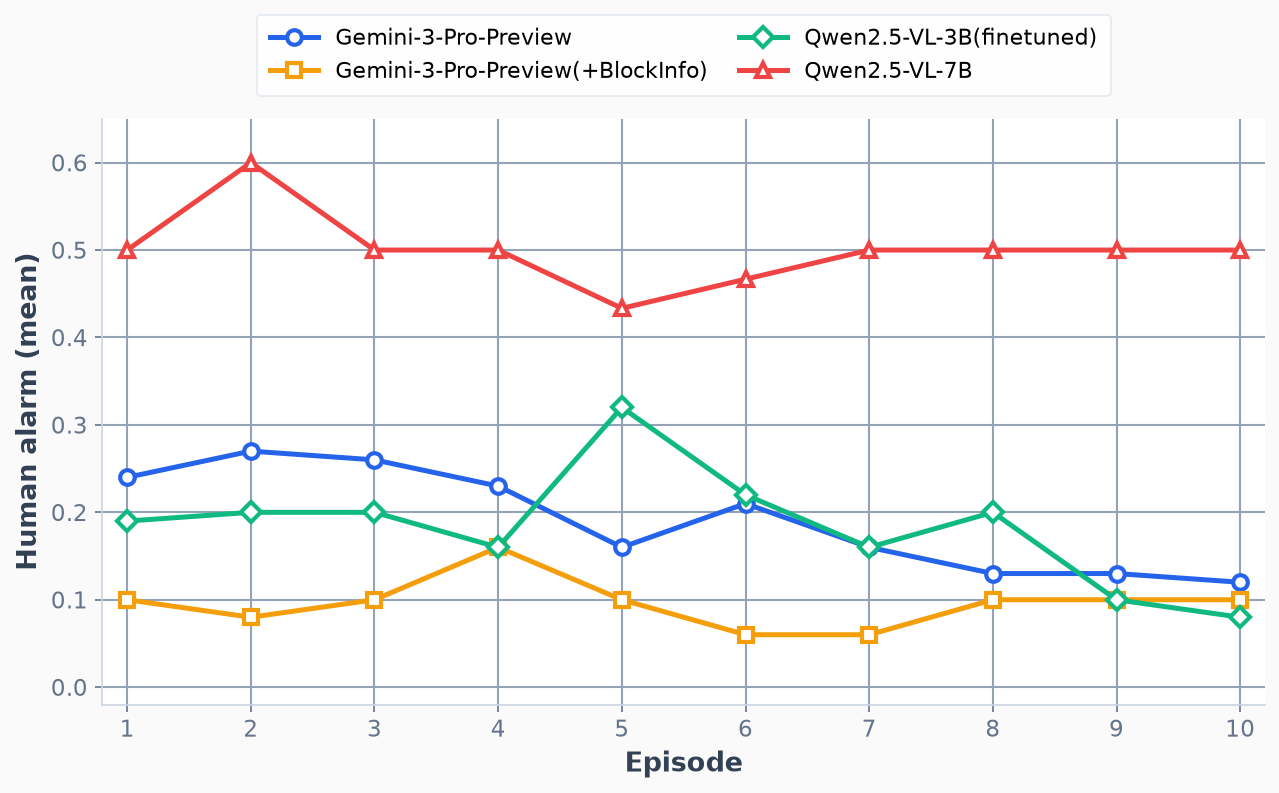}
    \caption{Mean Human Alarm across environments.}
    \label{fig:alarm}
\end{figure}

\textit{Human alarm rate.} Human alarm rate measures how often the critic triggers a safety or execution alert requiring human attention (Fig.~\ref{fig:alarm}). This signal reflects both error detection reliability and safety awareness during task execution.

Fine-tuned Qwen2.5-VL-3B critic shows a gradual decrease in alarm frequency across episodes. As policy quality improves and execution errors become less frequent, the number of triggered alarms correspondingly declines. Since the critic remains frozen, the decreasing alarm rate reflects improvements in the Behavior Tree and a reduction in observable execution failures, rather than adaptation of the critic itself.

In contrast, Qwen2.5-VL-7B produces a higher and more variable alarm rate. The instability suggests less consistent confidence and alarm behavior and less consistent distinction between critical failures and minor execution imperfections.

The Gemini-3-Pro-Preview critic demonstrates a more stable alarm profile, with relatively high alarm rates during early episodes followed by a steady reduction as the policy improves. This behavior is desirable, as it reflects sensitivity to genuine failures during exploration while avoiding excessive intervention later.

The Gemini (+BlockInfo) configuration achieves the most balanced alarm behavior. Early alarms correctly capture orientation and placement errors, while later episodes show a significant reduction in unnecessary alerts. The structured perceptual input improves both detection consistency and safety-related alarm behavior.

\begin{figure}
    \centering
    \includegraphics[width=1\linewidth]{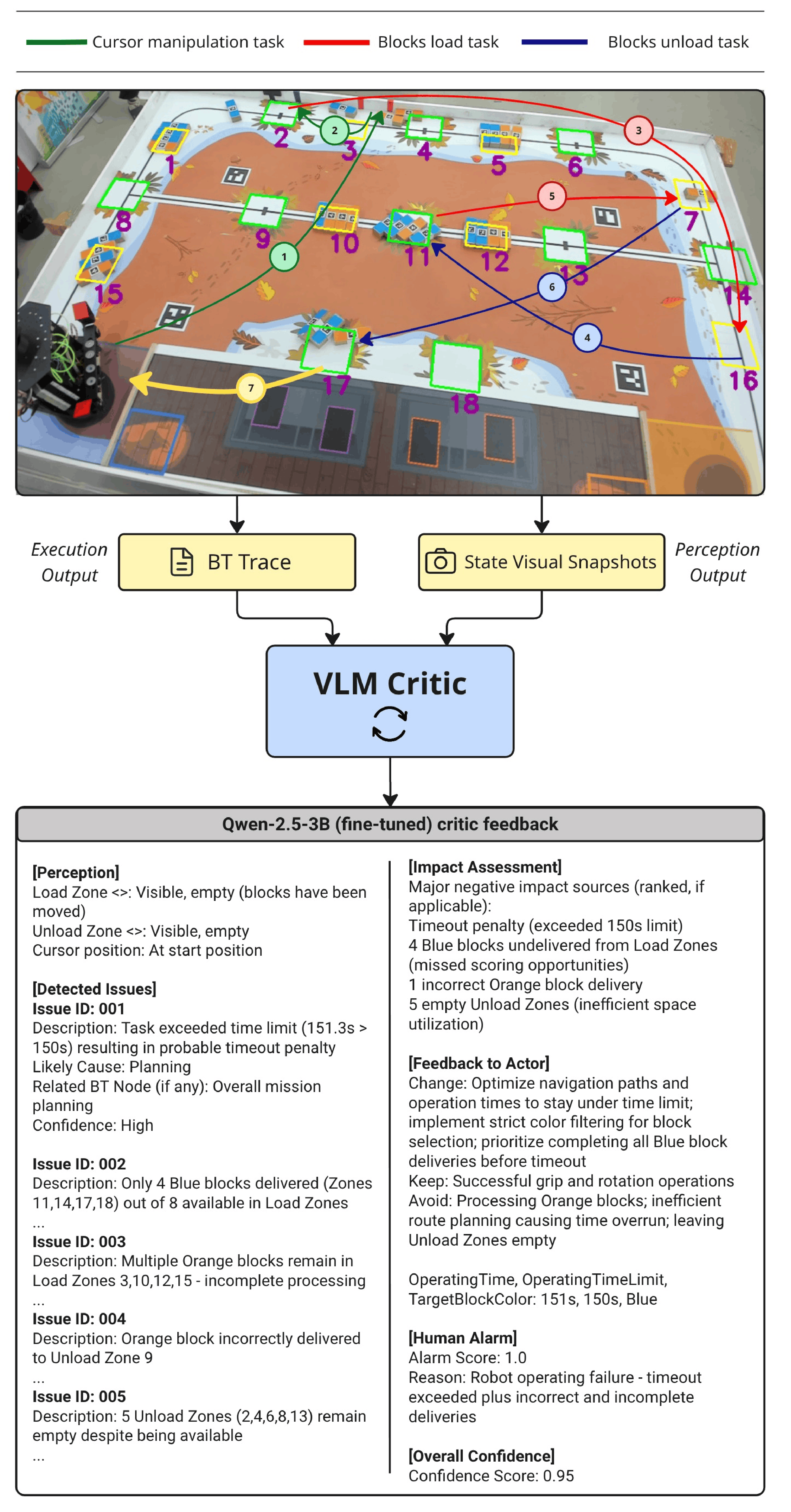}
    \caption{Episode of the Qwen2.5-VL-3B (fine-tuned) critic.}
    \vspace{-0.65cm}
    \label{fig:highlight}
\end{figure}

Particular attention is given to a representative episode of the Qwen2.5-VL-3B (fine-tuned) critic. In Fig.~\ref{fig:highlight}, we show the full robot trajectory during the task, together with the intermediate visual inputs processed by the VLM and the corresponding critic feedback. Based on the structured output of the fine-tuned critic, the system correctly identifies the key failure in the episode and, importantly, attributes it to the appropriate Impact Issue category and related BT node. This precise classification allows the actor to understand not only that an error occurred, but how it affected overall task planning and execution. Such impact-aware attribution is essential for targeted correction rather than global, unfocused replanning.

\begin{figure}
    \centering
    \includegraphics[width=1\linewidth]{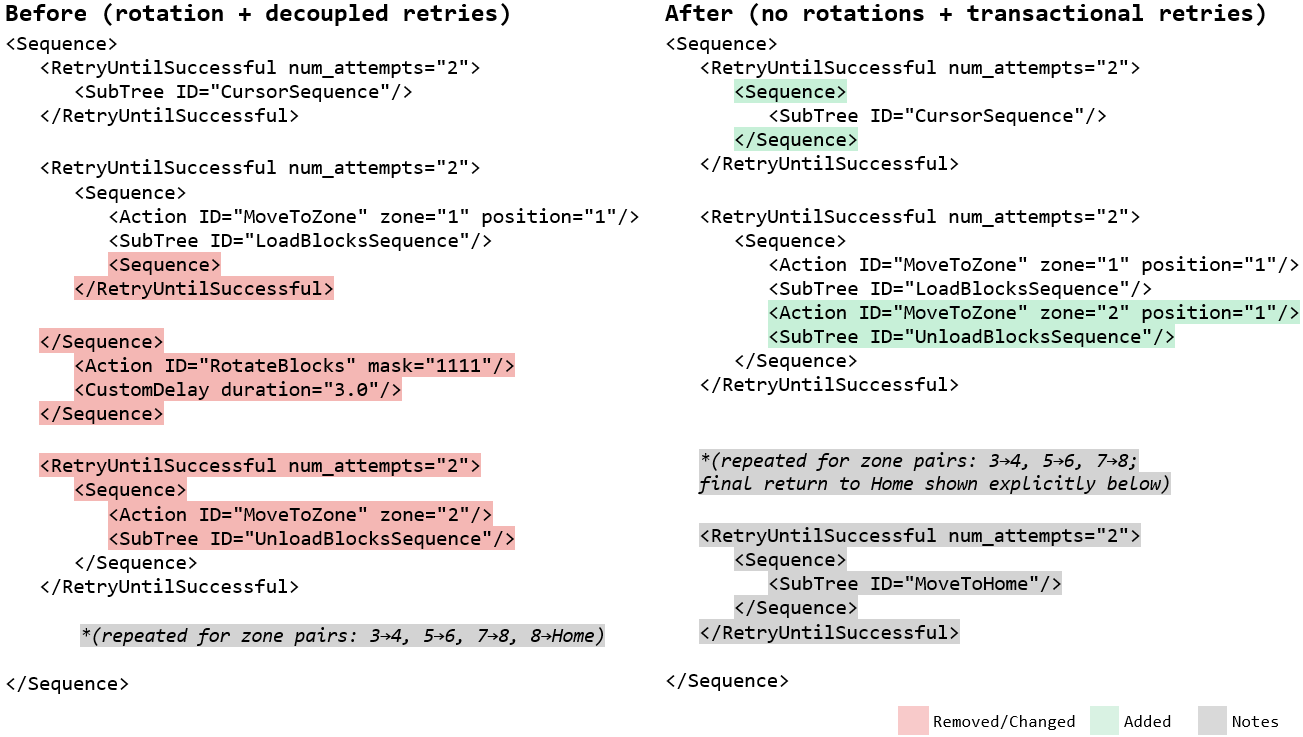}
    \caption{Qualitative BT evolution under perceptual uncertainty.}
    \vspace{0.25cm}
    \label{fig:case_rotation}
\end{figure}

\subsection{Qualitative BT Evolution Under Perceptual Uncertainty}

To complement the aggregate learning curves, Fig.~\ref{fig:case_rotation} illustrates a representative refinement step where the critic's perceptual grounding is insufficient for reliable color/orientation reasoning. In this episode, the critic reports empty unload zones and flags a generic execution failure, whereas the BT execution trace and the human-verified episode score $R$ indicate successful transport and placement. The actor therefore treats the critic output as a noisy hypothesis and grounds the update in the BT trace and the human-verified score $R$, applying a conservative and interpretable BT patch: (i) disabling the blind \texttt{RotateBlocks(mask=1111)} routine to avoid time overhead and accidental mis-rotations when color perception is unreliable; (ii) fixing an execution-engine edge case by wrapping \texttt{CursorSequence} inside a \texttt{Sequence} under \texttt{RetryUntilSuccessful}; and (iii) restructuring the policy into isolated transactional \texttt{Load--Transport--Unload} subtrees wrapped with \texttt{RetryUntilSuccessful}, preventing macro-retries across multiple zone pairs. This behavior also motivates our BlockInfo ablation, where reliable symbolic color/orientation estimates enable reintroducing targeted rotations and improve convergence.

\section{Discussion}

The experimental results indicate that critic alignment with the task domain plays a more significant role than raw model scale. Although Qwen2.5-VL-3B contains fewer parameters than the 7B variant, the fine-tuned 3B model demonstrates more stable learning dynamics and stronger improvement in $R$. This suggests that adaptation to structured execution feedback and alignment with Behavior Tree–level error categories are more important than generic visual capacity. The larger untuned model retains broader visual knowledge but lacks tight alignment with the symbolic planning abstraction, resulting in less consistent and less actionable critiques.

Gemini-3-Pro-Preview achieves stronger overall performance, which can be attributed to higher-quality visual grounding and more robust spatial reasoning. The task requires resolving fine-grained spatial relationships, such as identifying block orientation from colored faces, verifying full-batch completion in unload zones, and detecting shelf displacement along field boundaries. Small perceptual errors directly translate into discrete symbolic consequences at the planning level. In particular, accurate recognition of the movable shelf position along the boundary is critical, since even minor misalignment affects the validity of task completion. Gemini’s stronger multimodal reasoning enables more precise and consistent error detection, which in turn enables more effective Behavior Tree refinement. When additional symbolic block color information is provided, perceptual ambiguity is further reduced, shifting the burden from low-level visual inference to structured reasoning. This explains the increased stability and higher final performance of the Gemini (+BlockInfo) configuration.

The actor also adapts to the frozen critic's limitations. As illustrated in
Fig.~\ref{fig:case_rotation}, it reduces reliance on perception-sensitive
operations when color or zone evidence is unreliable. Thus, adaptation occurs
on the actor side: the critic defines the feedback signal, while the actor
adjusts the symbolic strategy accordingly.



Despite these strengths, the system’s performance remains fundamentally constrained by what the critic can observe and attribute within its limited temporal context. The current VersualRL formulation assumes that relevant faults are visible within the critic’s snapshot-based window and can be causally linked to recent BT events. Detecting latent faults with delayed consequences remains challenging, as errors may only manifest several steps later, outside the immediate observation context. Addressing this limitation likely requires longer-horizon temporal reasoning and more explicit trace-aware root-cause attribution.

A second limitation is that VersualRL inherits the critic's perceptual reliability. When the critic is insufficiently adapted, perceptual mismatches can yield incorrect causal hypotheses. As shown in Fig.~\ref{fig:case_rotation}, the actor can mitigate this by adopting conservative updates that reduce reliance on uncertain perceptual cues, but this may trade off optimality (e.g., disabling rotations that could be beneficial under reliable color observation). Improving critic grounding and incorporating explicit uncertainty handling or structured perceptual inputs (BlockInfo) are important directions.

A third limitation is the restricted scope of the experimental evaluation. The framework is validated on a single mobile manipulation platform and a structured warehouse-style task with a fixed library of Behavior Tree nodes. Although the results demonstrate closed-loop adaptation across multiple field configurations, further experiments are needed to assess generalization to unseen layouts, different robot embodiments, and tasks with richer manipulation and interaction dynamics.

Overall, these findings support the central premise of the proposed framework: effective task-level reinforcement in physical robotics does not rely on end-to-end differentiable policy learning, but rather on reliable, structured, and well-aligned verbal feedback grounded in observable execution outcomes. At the same time, they highlight that further improvements in critic grounding, temporal reasoning, and structured feedback design are key directions for advancing Verbal Reinforcement Learning in real-world robotic systems.

Future work will evaluate VersualRL on new robot embodiments, unseen layouts, and richer manipulation tasks; extend the critic with longer video context and trace-aware root-cause attribution; and investigate fully open-source critic configurations to improve reproducibility.

\section{Conclusion}
We presented VersualRL, a closed-loop verbal reinforcement learning framework with visual execution feedback for task-level policy refinement in mobile robotics. VersualRL addresses the problem of execution uncertainty in stochastic real-world settings by combining a VLM-based critic with an LLM-based actor. The critic generates structured natural-language feedback based on robot observations, allowing the actor to iteratively refine Behavior Trees without gradient-based optimization.

Key findings from experiments with physical robots include:

1. The human-verified score $R$ alone is insufficient under execution uncertainty, as it cannot distinguish execution (hardware) errors from planning errors. Structured feedback is essential for stable policy improvement.

2. The quality of causal attribution is more important than perceptual accuracy. A critic that correctly diagnoses the causes of failures ensures robust adaptation even with imperfect object detection. 

3. Structured perceptual inputs accelerate convergence, demonstrating that the main bottleneck of VersualRL is the perceptual foundation, not symbolic planning.

VersualRL enables interpretable, hardware-based policy adaptation in the real world—without simulators or gradients. Performance depends on the reliability of the critic and the design of the contextual cues, but this framework suggests a promising direction toward transparent robot learning that accounts for the gap between plan and reality.

\section*{Acknowledgements} 

Research reported in this publication was financially supported by the RSF grant No. 24-41-02039.

The authors would like to thank Igor Duchinskii, Nikita Kuzmin, Mariya Lezina and Georgii Demianchuk for their valuable assistance with data collection.

\bibliographystyle{IEEEtran}
\bibliography{bib}

\end{document}